\documentclass[journal]{IEEEtran}
\usepackage{array}
\usepackage{multicol}
\usepackage[caption=false,font=normalsize,labelfont=sf,textfont=sf]{subfig}
\usepackage{textcomp}
\usepackage{stfloats}
\usepackage{url}
\usepackage{verbatim}
\usepackage{graphicx}
\usepackage{tikz}
\usepackage{graphicx}
\usepackage{cite}
\usepackage{multirow}
\usepackage{makecell}
\usepackage[ruled,linesnumbered]{algorithm2e}
\usepackage{amsmath}
\usepackage{bm}
\usepackage{color}
\usepackage{graphicx}
\usepackage{amssymb}
\usepackage{url}
\usepackage{booktabs}
\usepackage{multirow}
\usepackage[normalem]{ulem}
\usepackage{ragged2e}
\usepackage{soul}
\usepackage{xcolor}
\usepackage{ulem}
\usepackage{colortbl}
\usepackage{booktabs}
\usepackage[normalem]{ulem}
\useunder{\uline}{\ul}{}
\usepackage[colorlinks,linkcolor=blue,citecolor=blue]{hyperref}
\usepackage{longtable}
\usepackage{relsize}
\usepackage{comment}
\usepackage{diagbox}
\newcolumntype{C}[1]{>{\centering\arraybackslash}p{#1}}
\newcolumntype{L}[1]{>{\raggedright\arraybackslash}p{#1}}

\begin{document}

\title{FH-Sampling: Frequency-aware Hypergraph-driven Oversampling for Imbalanced Time-Series Classification}

\author{
Yunrong~Hao,
Ruohao~Dai,
Zhen~Liu,
Guan~Wang,
Wenbin~Pei,
and Qiang~Zhang,~\IEEEmembership{Senior~Member,~IEEE}
\thanks{This work was supported in part by the National Key Research and Development Program of China under grant 2024YFA1012700, the National Natural Science Foundation of China under grants 62206041, 12371516 and U21A20491, and the NSFC-Liaoning Province United Foundation under grant U1908214, the 111 Project under grant D23006, the Liaoning Revitalization Talents Program under grant XLYC2008017, and China University Industry-University-Research Innovation Fund under grants 2022IT174, Natural Science Foundation of Liaoning Province under grant 2023-BSBA-030, and an Open Fund of National Engineering Laboratory for Big Data System Computing Technology under grant SZU-BDSC-OF2024-09. (Corresponding author: Wenbin~Pei.)
}
\thanks{Yunrong Hao and Wenbin Pei are with the School of Computer Science and Technology, Dalian University of Technology, Dalian 116024, China; Key Laboratory of Social Computing and Cognitive Intelligence (Dalian University of Technology), Ministry of Education, Dalian 116024, China (e-mail: haoyunrong@mail.dlut.edu.cn; peiwenbin@dlut.edu.cn).}%
\thanks{Ruohao Dai is with the School of Engineering and Computer Science, Victoria University of Wellington, PO Box 600, Wellington 6140, New Zealand (e-mails: dairuoh@myvuw.ac.nz).}%
\thanks{Zhen Liu is with the College of Computing and Data Science, Nanyang Technological University, Singapore (e-mail: zhen.liu@ntu.edu.sg).}%
\thanks{Guan Wang is with the School of Mechanics and Aerospace Engineering, Dalian University of Technology, Dalian 116024, China (e-mail: guanwang@dlut.edu.cn).}%
\thanks{Qiang Zhang is with the School of Computer Science and Technology, 
Dalian University of Technology, Dalian 116024, China; 
Key Laboratory of Social Computing and Cognitive Intelligence 
(Dalian University of Technology), Ministry of Education, Dalian 116024, China; 
and also with the National and Local Joint Engineering Laboratory of Computer Aided Design, 
Dalian University, Dalian 116622, China 
(e-mail: zhangq@dlut.edu.cn).}}

\maketitle

\markboth{Journal of \LaTeX\ Class Files,~Vol.~14, No.~8, August~2015}%
{Shell \MakeLowercase{\textit{et al.}}: Bare Demo of IEEEtran.cls for IEEE Journals}


\begin{abstract}
Imbalanced time-series data appears commonly in diverse applications,
such as medical diagnosis. When trained on imbalanced data, classifiers
tend to exhibit a bias toward the majority class, making it difficult
to identify minority-class instances accurately, despite their critical
importance. Sampling techniques have proven effective in rebalancing
imbalanced data. However, existing interpolation-based oversampling
methods typically rely on pairwise interpolation, limiting their ability
to capture high-order inter-sample relationships, and their inherent
smoothing effects tend to suppress high-frequency components,
compromising spectral fidelity. Therefore, we propose a
\textbf{F}requency-aware \textbf{H}ypergraph-driven \textbf{O}versampling
method (\textbf{FH-Sampling}) for imbalanced time-series classification.
In FH-Sampling, a hypergraph is first constructed in the original
time-series space to capture complex high-order inter-sample
relationships, and sampling ratios are adaptively assigned to mitigate
the impact of noisy samples.
The discrete wavelet transform decomposes each sample into frequency
components. Guided by the sample hypergraph, FH-Sampling combines
high-frequency components from related samples while retaining the
target sample's low-frequency component to help preserve important
frequency characteristics.
Experiments on 24 imbalanced datasets demonstrate the advantages
of FH-Sampling for imbalanced time-series classification. We further
apply FH-Sampling to aircraft trajectory generation. Experiments on
simulated aircraft trajectories show that FH-Sampling produces
trajectories that better match the real distribution, highlighting
its potential for practical applications.
\end{abstract}


\section{Introduction}

\IEEEPARstart{T}{ime}-series classification (TSC) aims to assign a class label to a sequence of observations collected over time~\cite{foumani2024deep,abdullayev2026enhancing}. It supports a wide range of applications, including medical diagnosis~\cite{Wang2020medical_diagnostics}, industrial fault detection~\cite{10028769}, and human activity recognition~\cite{jiang2015human}. However, real-world time-series data often exhibit class imbalance, with majority-class samples substantially outnumbering minority-class samples~\cite{gao2026imbalanced}. This can bias classifiers toward majority classes~\cite{chen2024survey}, hindering the recognition of minority-class instances that often represent critical conditions, such as physiological abnormalities~\cite{Wang2020medical_diagnostics} or equipment faults~\cite{LI2025110410}.

Sampling techniques effectively reduce model bias in imbalanced time-series classification (ITSC) by rebalancing the original data distribution~\cite{5128907,cao2013integrated}. Oversampling is a common sampling strategy that rebalances data by generating new samples from the minority class~\cite{chawla2002smote}. The most representative oversampling method is Synthetic Minority Over-sampling Technique (SMOTE)~\cite{chawla2002smote}. It generates synthetic samples by interpolating between minority-class samples and their same-class neighbors~\cite{chawla2002smote}. While standard SMOTE can be applied to ITSC, it fails to account for the temporal dependencies and dynamic patterns inherent in time-series data~\cite{zhao2022t}. To address this limitation, the Temporal-oriented Synthetic Minority Oversampling Technique (T-SMOTE) is introduced in~\cite{zhao2022t} to synthesize time-series data more effectively.

Although recent methods demonstrate effectiveness in synthesizing the minority class, they have the following major limitations:
\begin{itemize}
    \item The generation of minority-class samples typically relies on linear interpolation between two sequences, failing to capture the complex high-order inter-sample relationships that characterize real-world temporal patterns~\cite{zhao2026fuzzyhypergraph}. This may cause the generation of noisy instances~\cite {SAGLAM2025113906}, potentially shifting the classification boundary and limiting model generalization in complex ITSC tasks.
    
    \item The generation of minority-class samples neglects frequency-domain information, resulting in synthetic samples that lack authentic spectral characteristics essential for accurate discriminative patterns of time-series. However, in many domains such as medical diagnostics~\cite{zhang2022self} and industrial fault detection~\cite{10028769}, frequency components (e.g., heart rate variability or vibration frequencies) are essential for accurate classification. 
\end{itemize}

To address these limitations, we propose \textbf{F}requency-aware \textbf{H}ypergraph-driven Over\textbf{sampling} (\textbf{FH-Sampling}) for ITSC. To capture the high-order inter-sample relationships overlooked by interpolation between two sequences, FH-Sampling constructs a sample hypergraph. These high-order relationships guide sample synthesis, thereby reducing the generation of noisy samples and their potential impact on the classification boundary. Note that, even when high-order inter-sample relationships are considered, direct interpolation in the time domain may still suppress important frequency characteristics. FH-Sampling therefore performs sample synthesis in the wavelet coefficient space obtained through the discrete wavelet transform (DWT)~\cite{chaovalit2011discrete}. 
The main contributions of this study are summarized as follows:
\begin{itemize}
    \item We construct a sample hypergraph to capture high-order inter-sample relationships and adaptively allocate synthetic samples among minority-class instances. This allows generation to use information from multiple related samples, helping reduce the influence of noisy instances on sample synthesis.

    \item 
We develop a wavelet-domain synthesis mechanism that combines
high-frequency components of samples within a hyperedge while
retaining the target sample's low-frequency component. This
incorporates higher-order relationships among samples into data generation and
helps mitigate the loss of frequency information. 

    \item We conduct extensive comparisons across datasets and classifiers. These evaluations demonstrate the effectiveness of FH-Sampling, clarify the contributions of its components, and assess its computational efficiency. We further apply the proposed method to aircraft trajectory generation, evaluating generation quality on simulated trajectories to demonstrate its potential for practical applications.
\end{itemize}

The remainder of this paper is organized as follows. Section~\ref{background} introduces the background on hypergraphs and discrete wavelet decomposition, followed by a review of related work. Section~\ref{method} presents the proposed FH-Sampling method. Section~\ref{experimental design} describes the experimental setup, while Section~\ref{results and analysis} reports and analyzes the experimental results. Section~\ref{aircraft trajectory generation}
presents an application of FH-Sampling to aircraft trajectory generation. Finally, Section~\ref{conclusion} concludes the paper.

\section{Background}    
\label{background}

\subsection{Hypergraphs}

\begin{figure}[t]
    \centering
    \includegraphics[width=\columnwidth]{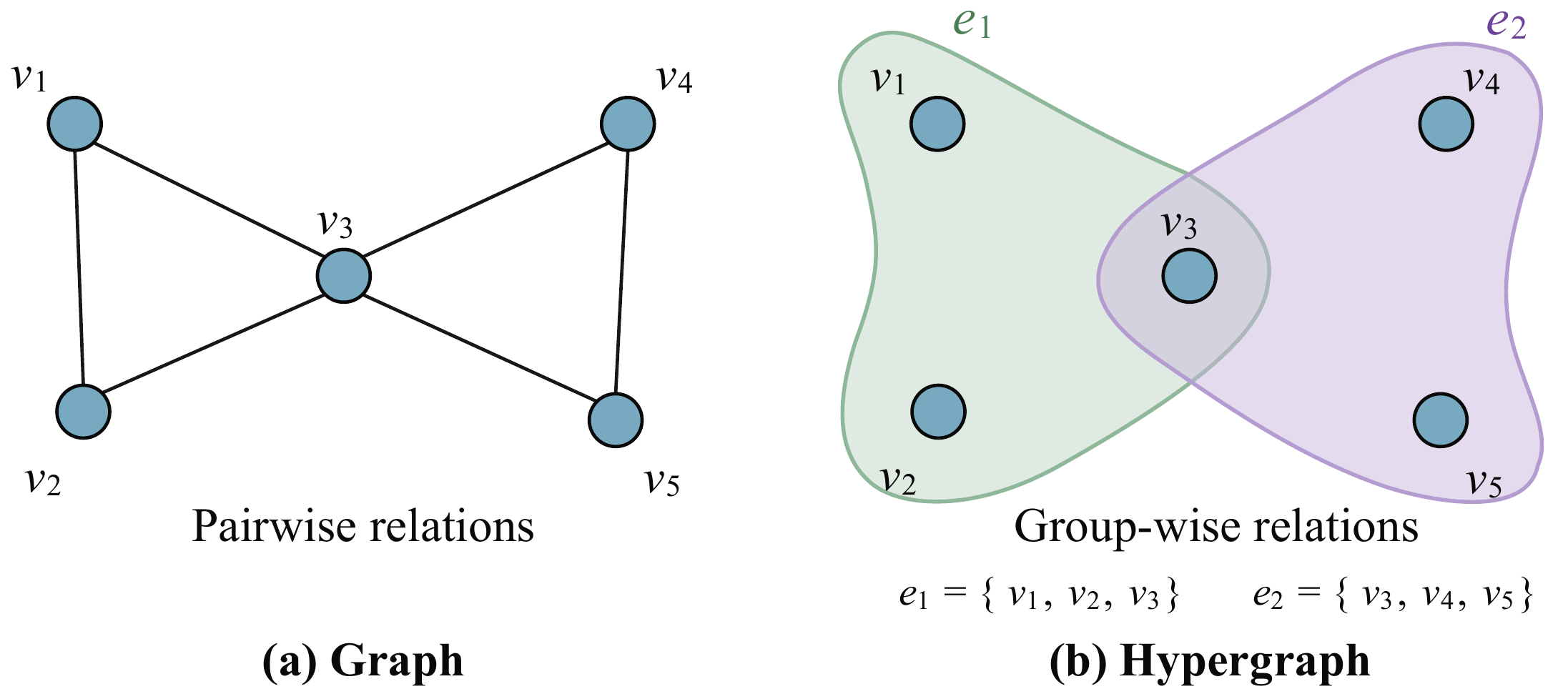}
    \caption{Comparison of an ordinary graph and a hypergraph. (a) Edges represent pairwise relationships. (b) Hyperedges represent group-wise relationships, with vertex $v_3$ shared by $e_1$ and $e_2$.}
    \label{fig:hypergraphs}
\end{figure}

A hypergraph generalizes a conventional graph by allowing a hyperedge to connect multiple vertices, thereby explicitly representing high-order relationships~\cite{9264674}. As illustrated in Fig.~\ref{fig:hypergraphs}, an ordinary graph captures pairwise relationships through edges, whereas a hypergraph connects multiple vertices through a single hyperedge. In Fig.~\ref{fig:hypergraphs}(b), the hyperedges are $e_1=\{v_1,v_2,v_3\}$ and $e_2=\{v_3,v_4,v_5\}$.

Formally, a weighted hypergraph is denoted by
$\mathcal{G}=(\mathcal{V},\mathcal{E},\mathcal{W})$,
where $\mathcal{V}$ is the vertex set,
$\mathcal{E}$ is the hyperedge set, and
$\mathcal{W}$ is the hyperedge weights set. Each hyperedge $e$ is a nonempty subset of $\mathcal{V}$.

The membership of vertices in hyperedges is encoded by an incidence matrix
$\mathbf{H}\in\{0,1\}^{|\mathcal{V}|\times|\mathcal{E}|}$~\cite{feng2019hypergraph}:
\begin{equation}
    H(v,e)=
    \begin{cases}
        1, & v\in e,\\
        0, & v\notin e.
    \end{cases}
    \label{eq:hypergraph_incidence}
\end{equation}
Each row corresponds to a vertex, and each column corresponds to a hyperedge. A column therefore identifies all vertices belonging to a particular hyperedge, while a row identifies all hyperedges containing a particular vertex.

Vertex and hyperedge degrees summarize two aspects of this connectivity. The weighted degree of a vertex $v$ is defined as
\begin{equation}
    d(v)=\sum_{e\in\mathcal{E}}w(e)H(v,e),
    \label{eq:vertex_degree}
\end{equation}
which accumulates the weights of all hyperedges containing $v$.

The degree of a hyperedge $e$ is defined as
\begin{equation}
    \delta(e)=\sum_{v\in\mathcal{V}}H(v,e)=|e|,
    \label{eq:hyperedge_degree}
\end{equation}
which gives the number of vertices it contains.

\subsection{Discrete Wavelet Transform}

The DWT decomposes a time series
into approximation and detail coefficients at multiple
scales~\cite{chaovalit2011discrete}. Approximation coefficients
describe lower-frequency structure, while detail coefficients
capture variations in the corresponding higher-frequency bands. In a single-level DWT, a discrete time-series $\mathcal{X}_i$ is filtered by a low-pass filter $g$ and a high-pass filter $h$~\cite{chaovalit2011discrete}, together with downsampling. This produces a low-frequency component $A_i^1$ and a high-frequency component $D_i^1$, which can be mathematically expressed as:
\begin{align}
  A_i^1[n] &= \sum_k x_i^{2n-k} \cdot g[k]; \label{eq:approximation}\\
  D_i^1[n] &= \sum_k x_i^{2n-k} \cdot h[k]. \label{eq:detail}
\end{align}

Simplified as:
\begin{equation}
(A_i^1,D_i^1)
=\operatorname{DWT}(\mathcal{X}_i).
\label{DWT}
\end{equation}

Multi-level DWT iteratively decomposes the low-frequency component of the sequence~\cite{rhif2019wavelet}, expressed as:
\begin{equation}
(A_i^{j+1},D_i^{j+1})
=\operatorname{DWT}(A_i^j).
\label{MDWT}
\end{equation}

The inverse discrete wavelet transform (IDWT) reconstructs the original sequence by upsampling and filtering the low-frequency and high-frequency components:
\begin{equation}
\mathcal{X}_i=\operatorname{IDWT}(A_i^1,D_i^1).
\label{IDWT}
\end{equation}

Similarly, IDWT can also be used to restore the decomposed components to the low-frequency components of the previous level:
\begin{equation}
A_i^j=\operatorname{IDWT}(A_i^{j+1},D_i^{j+1}).
\label{MIDWT}
\end{equation}

\subsection{Related Work}

\subsubsection{Hypergraphs for Time-Series Modeling}
Recent studies~\cite{shang2024ada,tian2025hypermixer,shang2024mshyper,jiang2025novel} investigate the use of hypergraphs for time-series modeling, demonstrating their effectiveness in capturing high-order relationships among channels or temporal segments. HyperMixer~\cite{tian2025hypermixer} introduces a dynamic hypergraph to capture group-wise channel interactions and temporal patterns for long-term multivariate forecasting, while the Multi-Scale Hypergraph Transformer (MSHyper)~\cite{shang2024mshyper} constructs hypergraphs based on temporal proximity to capture multi-scale high-order interactions among temporal patterns. 
For TSC, the Mamba-hypergraph-enhanced
time-frequency fusion network (MH-TFFN)~\cite{jiang2025novel} represents segmented time-series as hypergraph nodes to learn temporal dependencies and dynamic characteristics. 
However, these methods primarily model relationships among channels or temporal segments, instead of capturing high-order relationships among time-series samples.

\subsubsection{Frequency-Domain Modeling for Time-Series Classification}
The Fourier transform is commonly used in frequency-domain modeling methods for TSC. Previous studies~\cite{ma2019attnsense,zhang2022self} employ the discrete Fourier transform to convert signals from the time domain to the frequency domain, thereby identifying periodic patterns and spectral components. However, it suffers from fundamental limitations in local feature extraction due to its inherent global signal representation characteristics. In contrast, DWT~\cite{wang2023wavelet,yang2025wavelet} uses multiscale decomposition to capture both frequency information and local temporal features.

\subsubsection{Oversampling for Imbalanced Time-Series Classification}

Oversampling is a fundamental sampling technique for ITSC that generates synthetic minority-class samples before model training.

The representative oversampling method, SMOTE~\cite{chawla2002smote}, is designed for general imbalanced data, ignoring temporal dependencies in time-series data. The dynamic time warping (DTW)-based oversampling method (DTW-SMOTE)~\cite{yang2021time} leverages DTW-based neighbor selection to mitigate this issue.
Alternatively, covariance estimation and Gaussian modeling are employed to capture global distributional structures~\cite{cao2011spo,zhu2022minority}. However, these methods do not explicitly model local temporal patterns during sample generation.
In contrast, T-SMOTE~\cite{zhao2022t} interpolates between critical subsequences and their temporal neighbors near decision boundaries, effectively preserving temporal dependencies while improving minority-class recognition. However, its pairwise interpolation mechanism neither explicitly captures high-order relationships among multiple samples nor accounts for frequency-domain characteristics during synthesis. As a result, the generated samples may miss patterns shared by multiple minority-class samples and lose fine-scale details through interpolation. Therefore, in this work, we investigate how high-order relationships among time-series samples can guide minority-class sample synthesis while preserving frequency-domain characteristics.


\section{Proposed Sampling Method: FH-Sampling}
\label{method}

\subsection{Problem Definition}

Let $\mathcal{D} = \{(\mathcal{X}_i,\mathcal{Y}_i)\}_{i=1}^{N}$ be a labeled time-series training dataset, where $N$ denotes the total number of samples. Each univariate time series $\mathcal{X}_i=(x_i^1,x_i^2,\ldots,x_i^T)$ is represented as a sequence of length $T$. The corresponding label is $\mathcal{Y}_i\in\{1,2,\ldots,C\}$, where $C$ denotes the number of classes.

Let $n_c$ denote the number of training samples in class $c$, and let $n_{\max}=\max_{c\in\{1,\ldots,C\}}n_c$. Classes with fewer than $n_{\max}$ samples are treated as minority classes and require oversampling. Let $L_{\text{pos}}=\{c:n_c<n_{\max}\}$ and $L_{\text{neg}}=\{1,2,\ldots,C\}\setminus L_{\text{pos}}$ denote the label sets for minority and majority classes, respectively. These sets are used to distinguish classes for sampling; the original $C$-class labels are retained for classification.

The primary objective of FH-Sampling is to generate a labeled synthetic sample set $\mathcal{D}_{\text{syn}}$ such that $N_c=n_{\max}$ for every $c\in\{1,\ldots,C\}$, where $N_c$ denotes the cardinality of class $c$ in the dataset $\mathcal{D}\cup\mathcal{D}_{\text{syn}}$. Thus, $n_{\max}-n_c$ synthetic samples are generated for each minority class, while the test set remains unchanged.

\subsection{The Overview of FH-Sampling}

\begin{figure*}[!t]
\centering
\includegraphics[width=1\textwidth]{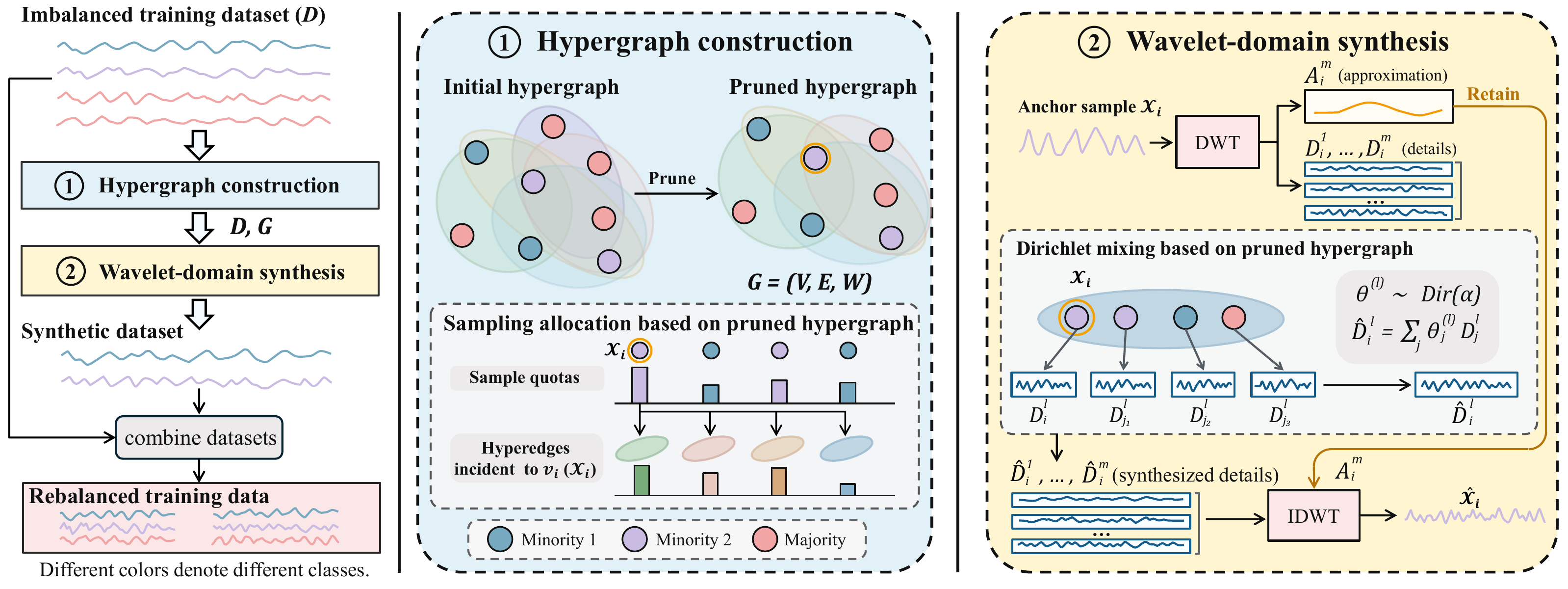}
\caption{The overview of FH-Sampling.}
\label{framework}
\end{figure*}

The flowchart of FH-Sampling is shown in Fig.~\ref{framework}.
FH-Sampling has two major processes:
(i) hypergraph construction;
(ii) wavelet-domain synthesis.

For the former, we construct a hypergraph over all time-series samples
to capture high-order relationships among them. A hyperedge pruning
mechanism is then applied to remove redundant or low-quality hyperedges.
The degree of each vertex in the pruned hypergraph is used to determine
the synthesis ratio for each minority-class sample. The synthetic samples are then further allocated to the associated hyperedges based on their weights.

For the sample synthesis process, the wavelet transform is first employed to decompose each time-series sample into low-frequency and high-frequency components. The high-frequency
components of a synthetic sample are generated by combining the
corresponding components of samples within a hyperedge.
Because low-frequency components reflect the global trend and structure
of the time series~\cite{Zhang2023Flight}, we directly retain the
low-frequency part of $\mathcal{X}_i$ as that of the synthetic sample.
Finally, IDWT is applied to reconstruct the
synthetic components back to the original time-series space.
The pseudocode of FH-Sampling is presented in
\textbf{Appendix A.1 of Supplementary Materials}.
Note that FH-Sampling, as a model-agnostic sampling method, can be
used to assist any classifier to mitigate the issue of class imbalance
in ITSC tasks.

\subsection{Hypergraph Construction and Allocation of the Synthesis Ratio}

\paragraph{Hypergraph Construction}
The time-series samples in a dataset are modeled by a hypergraph $\mathcal{G}$, where each vertex represents a sample. The vertex set is defined as $\mathcal{V}=\{v_1,v_2,\ldots,v_N\}$, where each vertex $v_i$ is a time-series sample $\mathcal{X}_i$ and $N$ denotes the total number of samples. The hyperedge set is denoted as $\mathcal{E}=\{e_1,e_2,\ldots,e_{|\mathcal{E}|}\}$, where $|\mathcal{E}|$ is the total number of hyperedges.

To construct the hypergraph $\mathcal{G}$, we adopt a $k$-nearest neighbors (KNN)-based strategy that captures the relationships among time-series samples based on their distances in the feature space~\cite{Huang2009Video}. Specifically, for each vertex $v_i$, we identify its $k$ nearest neighbors using KNN, forming a hyperedge $e_i$ that includes $v_i$ and its neighbors. This process is repeated for all vertices, thereby generating a total of $N$ hyperedges.

For each hyperedge $e\in\mathcal{E}$, its weight $w(e)$ is calculated as follows:
\begin{equation}
w(e)=
\sum_{v_i,v_j\in e}
\exp\left(
-\frac{d(v_i,v_j)^2}{\sigma^2}
\right),
\label{weight_edge}
\end{equation}
where $d(v_i,v_j)$ denotes the distance between vertex $v_i$ and $v_j$, and $\sigma$ is the median of all pairwise distances between vertices.

We use the incidence matrix $\mathbf{H}$ to store the relationships between vertices and hyperedges, defined as:
\begin{equation}
\mathbf{H}(v,e)=
\begin{cases}
1, & \text{if } v\in e,\\
0, & \text{otherwise}.
\end{cases}
\label{incidence_matrix}
\end{equation}

The degree of a vertex largely reflects its importance in the hypergraph. In this study, the degree of a vertex $v$ is defined as:
\begin{equation}
d(v)=
\sum_{e\in\mathcal{E}}
w(e)\mathbf{H}(v,e),
\label{degree_vertex}
\end{equation}
where $w(e)$ is computed according to equation (\ref{weight_edge}).

Moreover, redundant hyperedges should be removed to mitigate the impact of noisy samples. Therefore, we define the hyperedge quality $Q(e)$ as:
\begin{equation}
Q(e)=
1-
\frac{
\sum_{v_i\in e}
\mathbb{I}\bigl(\mathcal{Y}(v_i)\in L_{\text{neg}}\bigr)
}{
k+1
},
\label{edge_quality}
\end{equation}
where $\mathbb{I}(\cdot)$ is the indicator function, returning 1 if the condition is satisfied and 0 otherwise. $\mathcal{Y}(v_i)$ represents the class label of vertex $v_i$, and $L_{\text{neg}}$ is the label set for the negative classes.

To reduce the influence of noisy samples on vertex degrees, we introduce a quality threshold $\theta$, and remove any hyperedge $e$ if its quality $Q(e)<\theta$. Moreover, since intra-class easy samples typically have neighbors from the same class, we remove hyperedges containing only same-label vertices, to reduce the impact of such redundant intra-class connections. 
After pruning, we recompute the vertex degrees using the retained
hyperedges for the subsequent synthesis ratio allocation.

\paragraph{Allocation of the Synthesis Ratio}
For each sample $\mathcal{X}_i$ with label $\mathcal{Y}_i\in L_{\text{pos}}$, a synthesis ratio $r_i$ is assigned based on the degree of its corresponding vertex in the pruned hypergraph $\mathcal{G}$. The mathematical definition is as follows:
\begin{equation}
r_i=
\frac{
d(v_i)
}{
\sum_{v_j\in\mathcal{V},\;
\mathcal{Y}(v_j)=\mathcal{Y}(v_i)}
d(v_j)
},
\label{syn_ratio}
\end{equation}
where $v_j$ stands for sample $\mathcal{X}_j$, whose label is the same as that of $\mathcal{X}_i$ (including $\mathcal{X}_i$ itself). 

The target number of synthetic samples assigned to $\mathcal{X}_i$ is calculated as follows:
\begin{equation}
\tilde{n}_i=(n_{\max}-n_{\mathcal{Y}_i})r_i,
\label{syn_number}
\end{equation}
where $n_{\max}$ is the largest class size in the original training set, and $n_{\mathcal{Y}_i}$ is the number of original training samples in the class of $\mathcal{X}_i$. This ensures that each minority class receives exactly $n_{\max}-n_{\mathcal{Y}_i}$ synthetic samples.

In a hypergraph, a vertex $v_i$ may belong to multiple hyperedges.
The diversity of synthetic data is improved by interpolating between
sample $\mathcal{X}_i$ and its neighbors from distinct hyperedges.
After determining the number of synthetic samples assigned to
$\mathcal{X}_i$, we further use the hyperedge weights to determine
the proportion of synthetic samples generated within each associated
hyperedge, calculated as follows:

\begin{equation}
p(e_j)
=
\frac{w(e_j)}
{\sum_{e_\gamma\in\partial v_i}w(e_\gamma)},
\label{edge_ratio}
\end{equation}
where $e_\gamma\in\partial v_i$ indicates that the hyperedge
$e_\gamma$ is associated with the vertex $v_i$.

The target number of synthetic samples generated from
$\mathcal{X}_i$ within hyperedge $e_j$ is then calculated as:
\begin{equation}
\tilde n_{i,e_j}=\tilde n_i\,p(e_j),
\label{edge_syn_number}
\end{equation}
where $\tilde n_i$ is the target number of synthetic samples
assigned to $\mathcal{X}_i$.

\subsection{Wavelet-domain Synthesis}
The wavelet-domain synthesis involves the following two steps:
\paragraph{Wavelet Decomposition}
Directly generating samples in the original time-series space makes
it difficult to simultaneously capture both global trends and local
variations. The wavelet transform decomposes time-series data into
multi-scale components, effectively separating dynamics between
frequency levels~\cite{yang2023waveform,zhang2025waveletmixer}.
We therefore apply DWT to each sample before synthesis, allowing
its overall trend and local details to be treated separately.

As illustrated in Fig.~\ref{framework}, DWT decomposes each sample
$\mathcal{X}_i$ into a low-frequency component $A_i^m$ and a set of
high-frequency components $\{D_i^1,D_i^2,\ldots,D_i^m\}$, where
$m$ denotes the decomposition level. The low-frequency component
captures the overall trend of the sequence, while the high-frequency
components describe local variations at different scales.

\paragraph{Generation of Synthetic Samples}

After allocating the synthesis ratios as described above, we generate synthetic samples for each minority-class sample within its associated hyperedges. The high-frequency components are synthesized by combining
the corresponding components of samples within a hyperedge, while
the target sample's low-frequency component is retained.

We use the Dirichlet distribution~\cite{Baddar2025Efficient} to
generate interpolation weights for combining the high-frequency
coefficients. It produces non-negative weights that sum to 1 and
is therefore well-suited for modeling normalized weight vectors.
Specifically, the probability density function of the Dirichlet
distribution is given by:
\begin{equation}
\operatorname{Dir}
(\boldsymbol{\theta};\boldsymbol{\alpha})
=
\frac{1}{B(\boldsymbol{\alpha})}
\prod_{i=1}^{k+1}
\theta_i^{\alpha_i-1},
\label{dir}
\end{equation}
where
$\boldsymbol{\theta}=(\theta_1,\ldots,\theta_{k+1})
\in\mathbb{R}^{1\times(k+1)}$
denotes the generated weight vector satisfying
$\sum_{i=1}^{k+1}\theta_i=1$ and $\theta_i\geq0$.
The parameter vector
$\boldsymbol{\alpha}=(\alpha_1,\ldots,\alpha_{k+1})
\in\mathbb{R}^{1\times(k+1)}$
controls the shape of the distribution, and $k$ stands for the number
of neighbors in KNN. The normalization constant
$B(\boldsymbol{\alpha})$ is the multivariate Beta function, which
ensures that the distribution integrates to one.

The expected value of each $\theta_i$ is:
\begin{equation}
\mathbb{E}[\theta_i]
=
\frac{\alpha_i}{\sum_{j=1}^{k+1}\alpha_j},
\label{expectation}
\end{equation}
which indicates that the relative magnitude of $\alpha_i$ determines
the expected proportion assigned to each vertex in the generated
vector.

At each decomposition level, we draw an interpolation weight vector
$\boldsymbol{\theta}$ from the Dirichlet distribution.
For the $j$-th level high-frequency component of sample
$\mathcal{X}_i$, the corresponding synthetic component
$\hat{D}_i^j$ is computed as follows:
\begin{equation}
\hat{D}_i^j
=
\boldsymbol{\theta}e^j,
\label{syn}
\end{equation}
where $e^j\in\mathbb{R}^{(k+1)\times d_j}$ is the matrix whose rows
contain the $j$-th level detail coefficients of the samples in the
corresponding time-domain hyperedge containing $v_i$.
Here, $d_j$ denotes the coefficient length at this level.

Finally, the synthetic sample is reconstructed by retaining the
original low-frequency component $A_i^m$ and combining it with the
synthesized high-frequency components through multilevel IDWT:
\begin{equation}
\hat{\mathcal{X}}_i
=
\operatorname{IDWT}^{1}
\left(
\ldots
\operatorname{IDWT}^{m}
(A_i^m,\hat{D}_i^m),
\ldots,\hat{D}_i^1
\right).
\label{syn_sample}
\end{equation}

\section{Experimental Design}
\label{experimental design}
\subsection{Datasets}
To evaluate the proposed oversampling method, we use imbalanced time-series datasets from the UCR archive~\cite{dau2019ucr}. The selected 24 imbalanced datasets include real-world time series from medical diagnosis, industrial inspection, contour recognition, motion recognition and other scenarios. The datasets use the predefined training and test splits from the UCR archive.
We quantify class imbalance using the maximum imbalance ratio (MIR), defined
as the ratio of the largest to the smallest class size in the training set. The MIR of these datasets ranges from 1.03 to 33.83, covering diverse imbalance levels. For detailed information on these datasets, please refer to \textbf{Appendix B}.

\subsection{Baselines}
We compare FH-Sampling with six general-purpose oversampling methods: SMOTE~\cite{chawla2002smote}, Borderline-SMOTE1 (B1-SMOTE) and Borderline-SMOTE2 (B2-SMOTE)~\cite{han2005borderline}, SVM-SMOTE~\cite{nguyen2011borderline}, FE-SMOTE~\cite{chen2023new}, and H-SMOTE~\cite{LIAW2025121193}, as well as seven oversampling methods specifically designed for time-series data: DTW-SMOTE~\cite{yang2021time}, OHIT~\cite{zhu2022minority}, T-SMOTE~\cite{zhao2022t}, INOS~\cite{cao2013integrated}, BFGAN~\cite{bfgan2022}, CSMOTE~\cite{Liu2021CSmote}, and CFAMG~\cite{cfamg2025}. 
For detailed information on these baseline methods, please refer to \textbf{Appendix C}.

Note that FH-Sampling is a model-agnostic sampling method, which can be used with any classifier to mitigate the issue of class imbalance in TSC tasks. Therefore, in the experiments, in addition to the KNN classifiers, we also employ two widely used classifiers, namely support vector machines (SVM)~\cite{cortes1995support} and long short-term memory (LSTM) networks~\cite{hochreiter1997long}, as downstream backbone models to evaluate the effectiveness of the sampling methods in generating high-quality data for mitigating the issue of class imbalance.

\subsection{Experimental Settings}

\begin{table*}[t]
\centering
\caption{Comparison using the KNN classifier on the 24 datasets}
\label{tab:comparison_1nn}
\begingroup\footnotesize
\setlength{\tabcolsep}{4pt}
\renewcommand{\arraystretch}{1.2}
\resizebox{\textwidth}{!}{%
\begin{tabular}{@{}c*{12}{c}@{}}
\toprule[0.9pt]
& \multicolumn{3}{c}{BA} & \multicolumn{3}{c}{F1-Score} & \multicolumn{3}{c}{G-Mean} & \multicolumn{3}{c}{AUC} \\
\midrule[0.35pt]
Method & Avg.Value & Avg.Rank & $p$-value & Avg.Value & Avg.Rank & $p$-value & Avg.Value & Avg.Rank & $p$-value & Avg.Value & Avg.Rank & $p$-value \\
\midrule[0.55pt]
None & 0.6290 & -- & -- & 0.6266 & -- & -- & 0.5670 & -- & -- & 0.7527 & -- & -- \\
\midrule[0.35pt]
SMOTE & 0.6353 & 7.188 & 1.26E-03 & 0.6268 & 7.479 & 2.95E-04 & 0.6104 & 6.562 & 2.95E-04 & 0.7555 & 7.396 & 7.35E-04 \\
B1-SMOTE & 0.6401 & 5.979 & 2.29E-02 & 0.6303 & 6.312 & 1.64E-02 & 0.6119 & 5.729 & 1.94E-02 & 0.7582 & 6.104 & 1.79E-02 \\
B2-SMOTE & 0.6348 & 7.479 & 2.88E-03 & 0.6282 & 7.479 & 2.60E-03 & 0.6080 & 6.979 & 3.50E-03 & 0.7553 & 7.604 & 4.68E-03 \\
SVM-SMOTE & 0.6354 & 7.229 & 6.50E-04 & 0.6275 & 7.271 & 3.73E-04 & 0.6032 & 7.312 & 1.50E-04 & 0.7559 & 7.229 & 1.24E-03 \\
FE-SMOTE & 0.6308 & 9.229 & 1.55E-03 & 0.6283 & 8.958 & 8.76E-04 & 0.5767 & 8.708 & 5.45E-04 & 0.7529 & 9.250 & 1.55E-03 \\
H-SMOTE & 0.6388 & 6.229 & 4.34E-03 & 0.6283 & 6.479 & 1.05E-02 & 0.6094 & 5.938 & 1.15E-02 & 0.7573 & 6.229 & 3.90E-03 \\
\midrule[0.35pt]
DTW-SMOTE & \underline{0.6410} & \underline{5.188} & 2.01E-01 & 0.6310 & \underline{5.938} & 5.16E-02 & \underline{0.6133} & \underline{5.396} & 1.07E-01 & \underline{0.7595} & \underline{5.354} & 1.21E-01 \\
OHIT & 0.6333 & 9.167 & 2.88E-03 & 0.6288 & 8.521 & 4.25E-03 & 0.5878 & 9.792 & 2.13E-03 & 0.7550 & 8.896 & 2.88E-03 \\
T-SMOTE & 0.6343 & 6.688 & 7.77E-02 & \underline{0.6322} & 6.000 & 1.07E-01 & 0.5753 & 7.271 & 3.86E-02 & 0.7566 & 6.375 & 6.36E-02 \\
INOS & 0.6363 & 8.896 & 5.68E-04 & 0.6270 & 9.062 & 1.09E-03 & 0.6050 & 8.271 & 2.78E-04 & 0.7558 & 8.896 & 1.40E-03 \\
BFGAN & 0.6256 & 8.583 & 1.26E-03 & 0.6201 & 8.688 & 1.73E-03 & 0.5657 & 8.896 & 5.26E-04 & 0.7500 & 8.604 & 1.73E-03 \\
CSMOTE & 0.6249 & 9.562 & 7.35E-04 & 0.6197 & 9.729 & 1.14E-03 & 0.5642 & 10.375 & 9.16E-04 & 0.7499 & 9.604 & 1.14E-03 \\
CFAMG & 0.6239 & 9.896 & 2.29E-04 & 0.6190 & 9.562 & 2.60E-04 & 0.5629 & 10.333 & 2.29E-04 & 0.7484 & 9.896 & 2.01E-04 \\
\midrule[0.35pt]
\textbf{FH-Sampling (Ours)} & \textbf{0.6499} & \textbf{3.688} & -- & \textbf{0.6403} & \textbf{3.521} & -- & \textbf{0.6249} & \textbf{3.438} & -- & \textbf{0.7650} & \textbf{3.562} & -- \\
\bottomrule[0.9pt]
\end{tabular}%
}
\endgroup
\end{table*}

In FH-Sampling, the number of nearest neighbors $k$, used for forming a hyperedge, is set to 5 by default, following the common practice in oversampling methods implemented in the \texttt{imbalanced-learn} library~\cite{lemaavztre2017imbalanced}. The hyperedge quality threshold $\theta$ is set to 0.5 by default, meaning that a hyperedge is discarded if the majority of its samples belong to the negative class.

The wavelet basis is selected for each dataset using only the training data and is kept fixed in the subsequent evaluations. Details of the selection procedure are provided in \textbf{Appendix~D}.
The wavelet decomposition level $m$ is set to the default configuration of the PyWavelets library, which is calculated as
$m = \left\lfloor \log_2\left(\frac{T}{l-1}\right) \right\rfloor$,
where $T$ denotes the length of the time series and $l$ denotes the decomposition filter length of the selected wavelet.

For the Dirichlet distribution used to generate the interpolation weights of high-frequency components, the parameter vector $\boldsymbol{\alpha}$ is determined by the labels of nodes within a hyperedge. We assign $\alpha_i = \mathrm{e}^{2}$ to the target node, $\alpha_j = \mathrm{e}$ to same-class neighbors, and $\alpha_{r} = 1$ to nodes from different classes.


\section{Results and Analysis}
\label{results and analysis}
\subsection{Main Results}
To evaluate the effectiveness of different oversampling methods, we apply each method to rebalance the training set of every dataset, while keeping the test set unchanged. The classifiers are trained on the rebalanced training sets and evaluated on the original test sets using four widely adopted metrics for imbalanced data: balanced accuracy (BA)~\cite{brodersen2010balanced}, F1-Score~\cite{manning2009introduction}, G-Mean~\cite{ferri2009experimental}, and AUC~\cite{fawcett2006introduction}.

As shown in Table~\ref{tab:comparison_1nn}, FH-Sampling is compared with 13 baseline oversampling methods on the 24 imbalanced time-series datasets using the KNN classifier. The results show that FH-Sampling achieves the best average performance across all four evaluation metrics. Specifically, in terms of BA, FH-Sampling achieves the highest average value of 0.6499 and the best average rank of 3.688. For F1-Score, FH-Sampling outperforms the other methods in both average value and average rank, achieving 0.6403 and 3.521, respectively. For G-Mean, FH-Sampling achieves an average value of 0.6249, which is 1.89\% higher than the second-best method. It also maintains the best average rank of 3.438. For AUC, FH-Sampling achieves an average score of 0.7650, 0.72\% higher than the second-best method, and the best average rank of 3.562. The details of the results are provided in \textbf{Appendix E}.

To assess the statistical significance of performance differences, we employ the Wilcoxon signed-rank test~\cite{demvsar2006statistical}. At the significance level of $p<0.05$, FH-Sampling significantly outperforms 11 of the 13 baseline methods in BA, F1-Score, and AUC, and 12 of the 13 baseline methods in G-Mean.

These results support the effectiveness of FH-Sampling in mitigating class imbalance and improving overall classification performance.

\subsection{FH-Sampling with SVM and LSTM Classifiers}

\begin{table*}[t]
\centering
\caption{Comparison using the SVM classifier on the 24 datasets}
\label{tab:comparison_svm}
\begingroup\footnotesize
\setlength{\tabcolsep}{4pt}
\renewcommand{\arraystretch}{1.2}
\resizebox{\textwidth}{!}{%
\begin{tabular}{@{}c*{12}{c}@{}}
\toprule[0.9pt]
& \multicolumn{3}{c}{BA} & \multicolumn{3}{c}{F1-Score} & \multicolumn{3}{c}{G-Mean} & \multicolumn{3}{c}{AUC} \\
\midrule[0.35pt]
Method & Avg.Value & Avg.Rank & $p$-value & Avg.Value & Avg.Rank & $p$-value & Avg.Value & Avg.Rank & $p$-value & Avg.Value & Avg.Rank & $p$-value \\
\midrule[0.55pt]
None & 0.6007 & -- & -- & 0.5912 & -- & -- & 0.4796 & -- & -- & 0.8349 & -- & -- \\
\midrule[0.35pt]
SMOTE & 0.6163 & \underline{6.708} & 3.86E-02 & 0.6027 & 6.833 & 1.81E-01 & 0.5613 & 6.271 & 3.58E-02 & \textbf{0.8434} & \underline{6.042} & 6.84E-01 \\
B1-SMOTE & 0.6146 & 7.667 & 2.85E-02 & 0.5992 & 7.417 & 6.80E-02 & 0.5552 & 7.521 & 1.77E-02 & 0.8401 & 6.875 & 3.60E-01 \\
B2-SMOTE & 0.6130 & 7.500 & 7.44E-03 & 0.5961 & 7.958 & 6.79E-03 & 0.5498 & 7.396 & 7.44E-03 & 0.8385 & 7.375 & 3.03E-01 \\
SVM-SMOTE & 0.6140 & 7.771 & 1.38E-02 & 0.5977 & 7.396 & 7.87E-02 & 0.5328 & 7.583 & 6.50E-04 & 0.8416 & 6.375 & 4.91E-01 \\
FE-SMOTE & 0.5926 & 9.375 & 9.58E-03 & 0.5851 & 8.792 & 2.69E-02 & 0.5375 & 8.375 & 5.33E-03 & 0.8282 & 9.000 & 9.58E-03 \\
H-SMOTE & 0.6123 & 7.146 & 3.86E-02 & 0.5984 & 6.771 & 1.91E-01 & 0.5528 & 7.438 & 1.92E-02 & 0.8351 & 7.208 & 5.46E-01 \\
\midrule[0.35pt]
DTW-SMOTE & 0.6169 & 7.000 & 1.63E-02 & \underline{0.6044} & \underline{6.458} & 1.91E-01 & \underline{0.5670} & \underline{5.979} & 1.07E-01 & \underline{0.8429} & \textbf{5.458} & 7.68E-01 \\
OHIT & \underline{0.6184} & 7.250 & 3.60E-01 & 0.5992 & 7.958 & 1.28E-01 & 0.5544 & 7.000 & 1.78E-01 & 0.8423 & 6.208 & 7.47E-01 \\
T-SMOTE & 0.5911 & 7.875 & 2.11E-02 & 0.5841 & 6.958 & 1.28E-01 & 0.4674 & 8.417 & 5.33E-03 & 0.8373 & 7.250 & 6.63E-01 \\
INOS & 0.6077 & 7.750 & 1.07E-01 & 0.5870 & 8.667 & 5.64E-02 & 0.5255 & 9.000 & 1.05E-02 & 0.8295 & 10.333 & 6.52E-03 \\
BFGAN & 0.6038 & 8.396 & 4.81E-03 & 0.5900 & 8.938 & 5.90E-03 & 0.5287 & 8.417 & 5.90E-03 & 0.8292 & 9.667 & 7.92E-03 \\
CSMOTE & 0.6034 & 8.417 & 7.92E-03 & 0.5888 & 9.167 & 6.52E-03 & 0.5288 & 8.938 & 3.73E-04 & 0.8346 & 9.167 & 2.49E-02 \\
CFAMG & 0.6121 & 7.542 & 5.26E-02 & 0.6010 & 6.625 & 3.90E-01 & 0.5397 & 8.188 & 1.79E-02 & 0.8378 & 7.958 & 2.52E-01 \\
\midrule[0.35pt]
\textbf{FH-Sampling (Ours)} & \textbf{0.6211} & \textbf{4.604} & -- & \textbf{0.6047} & \textbf{5.062} & -- & \textbf{0.5851} & \textbf{4.479} & -- & 0.8425 & 6.083 & -- \\
\bottomrule[0.9pt]
\end{tabular}%
}
\endgroup
\end{table*}

\begin{table*}[t]
\centering
\caption{Comparison using the LSTM classifier on the 24 datasets}
\label{tab:comparison_lstm}
\begingroup\footnotesize
\setlength{\tabcolsep}{4pt}
\renewcommand{\arraystretch}{1.2}
\resizebox{\textwidth}{!}{%
\begin{tabular}{@{}c*{12}{c}@{}}
\toprule[0.9pt]
& \multicolumn{3}{c}{BA} & \multicolumn{3}{c}{F1-Score} & \multicolumn{3}{c}{G-Mean} & \multicolumn{3}{c}{AUC} \\
\midrule[0.35pt]
Method & Avg.Value & Avg.Rank & $p$-value & Avg.Value & Avg.Rank & $p$-value & Avg.Value & Avg.Rank & $p$-value & Avg.Value & Avg.Rank & $p$-value \\
\midrule[0.55pt]
None & 0.4850 & -- & -- & 0.4468 & -- & -- & 0.3122 & -- & -- & 0.7552 & -- & -- \\
\midrule[0.35pt]
SMOTE & 0.5054 & 6.500 & 3.40E-02 & 0.4732 & 6.750 & 8.39E-02 & 0.4051 & \underline{5.333} & 9.70E-01 & 0.7596 & 5.583 & 7.26E-01 \\
B1-SMOTE & \underline{0.5197} & \underline{5.625} & 3.45E-01 & \underline{0.4853} & 5.875 & 4.06E-01 & \textbf{0.4128} & 5.521 & 8.48E-01 & \underline{0.7641} & 5.625 & 7.90E-01 \\
B2-SMOTE & 0.5033 & 7.042 & 3.22E-04 & 0.4632 & 7.792 & 1.50E-04 & 0.3968 & 7.167 & 2.77E-01 & 0.7481 & 7.792 & 2.25E-03 \\
SVM-SMOTE & 0.5121 & 6.042 & 2.49E-02 & 0.4820 & \underline{5.250} & 1.97E-01 & 0.3931 & 6.354 & 1.22E-01 & 0.7606 & \underline{5.000} & 4.73E-01 \\
FE-SMOTE & 0.4831 & 7.958 & 1.78E-03 & 0.4527 & 7.417 & 1.05E-02 & 0.3548 & 7.604 & 1.17E-02 & 0.7353 & 9.583 & 3.73E-04 \\
H-SMOTE & 0.4914 & 7.417 & 2.25E-03 & 0.4569 & 7.542 & 4.81E-03 & 0.3852 & 6.729 & 3.57E-01 & 0.7509 & 7.688 & 7.19E-03 \\
\midrule[0.35pt]
DTW-SMOTE & 0.5014 & 6.500 & 2.81E-03 & 0.4647 & 6.375 & 3.15E-02 & 0.3870 & 6.542 & 9.10E-02 & 0.7524 & 6.167 & 7.87E-02 \\
OHIT & 0.4931 & 7.417 & 5.68E-04 & 0.4554 & 7.750 & 4.30E-04 & 0.3762 & 7.417 & 1.96E-02 & 0.7516 & 6.917 & 1.01E-01 \\
T-SMOTE & 0.4631 & 10.375 & 1.67E-06 & 0.4306 & 10.750 & 6.56E-06 & 0.2848 & 10.729 & 5.93E-04 & 0.7367 & 10.104 & 1.27E-04 \\
INOS & 0.4825 & 8.667 & 1.24E-03 & 0.4430 & 8.833 & 1.09E-03 & 0.3257 & 9.292 & 5.11E-03 & 0.7250 & 8.375 & 1.05E-02 \\
BFGAN & 0.4853 & 9.583 & 3.66E-05 & 0.4503 & 9.333 & 7.63E-05 & 0.3581 & 8.833 & 2.23E-03 & 0.7429 & 9.500 & 6.39E-05 \\
CSMOTE & 0.4797 & 9.125 & 2.78E-04 & 0.4498 & 8.750 & 8.46E-04 & 0.3341 & 9.125 & 8.69E-03 & 0.7492 & 8.708 & 7.92E-03 \\
CFAMG & 0.4725 & 8.875 & 3.73E-04 & 0.4431 & 8.167 & 1.40E-03 & 0.3276 & 9.188 & 1.12E-03 & 0.7377 & 9.000 & 7.42E-04 \\
\midrule[0.35pt]
\textbf{FH-Sampling (Ours)} & \textbf{0.5212} & \textbf{3.875} & -- & \textbf{0.4889} & \textbf{4.417} & -- & \underline{0.4103} & \textbf{5.167} & -- & \textbf{0.7658} & \textbf{4.958} & -- \\
\bottomrule[0.9pt]
\end{tabular}%
}
\endgroup
\end{table*}

To further assess the applicability of FH-Sampling beyond the KNN classifier, we evaluate the sampling methods using SVM and LSTM on the same 24 datasets.

As shown in Table~\ref{tab:comparison_svm}, with the SVM classifier, FH-Sampling achieves the highest average values in BA, F1-Score, and G-Mean, reaching 0.6211, 0.6047, and 0.5851, respectively. It also achieves the best average ranks on these three metrics, with values of 4.604, 5.062, and 4.479. In particular, its average G-Mean is 3.18\% higher than that of the second-best method, DTW-SMOTE. For AUC, FH-Sampling achieves an average value of 0.8425, close to the highest value of 0.8434 obtained by SMOTE, and the third-best average rank of 6.083. Compared with the original imbalanced data without oversampling, FH-Sampling improves all four average metrics, particularly G-Mean, which increases from 0.4796 to 0.5851, representing a relative improvement of 22.00\%.

As shown in Table~\ref{tab:comparison_lstm}, with the LSTM classifier, FH-Sampling achieves the highest average BA, F1-Score, and AUC, with values of 0.5212, 0.4889, and 0.7658, respectively. It achieves the best average ranks in BA and F1-Score, reaching 3.875 and 4.417. Although its average G-Mean of 0.4103 is slightly lower than the 0.4128 achieved by B1-SMOTE, FH-Sampling obtains the best average G-Mean rank of 5.167. For AUC, it also achieves the best average rank of 4.958. It also improves all four average metrics over the original imbalanced data. The details of the results are provided in \textbf{Appendix F of Supplementary Materials}.

Taken together, the main experiments with KNN and the additional experiments with SVM and LSTM show that rebalancing the training data with FH-Sampling can benefit classifiers with different learning mechanisms. This illustrates the value of addressing class imbalance at the data level, where the same oversampling procedure can be used
with different classifiers.

\subsection{Ablation Analysis}

To investigate the contributions of different components, we evaluate four ablation variants: (1) w/o HG (without hypergraph-guided synthesis) replaces hypergraph-guided synthesis with same-class neighbor interpolation in the wavelet domain; (2) w/o DWT (without the discrete wavelet transform) retains the hypergraph-based procedure but synthesizes samples directly in the time domain; (3) w/o QP (without quality pruning) removes quality pruning; and (4) w/o DA (without degree-based allocation) replaces degree-based allocation with uniform allocation.

\begin{figure}[t]
\centering
\includegraphics[width=1\columnwidth]{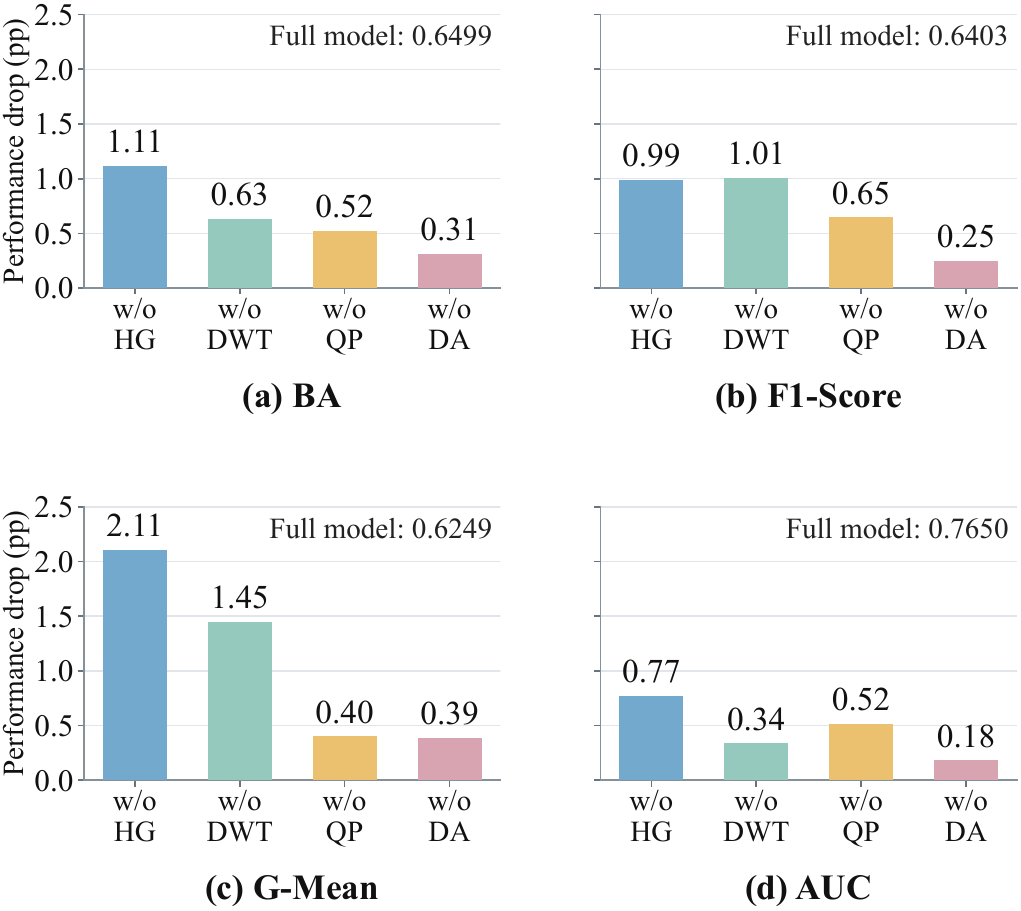}
\caption{Average performance decrease of each ablation variant relative to the complete FH-Sampling method on the 24 datasets. Decreases are measured in percentage points (pp).}
\label{fig:ablation_drop}
\end{figure}

As shown in Fig.~\ref{fig:ablation_drop}, removing any of the investigated components lowers the average performance across all four metrics. Removing hypergraph-guided synthesis causes the largest decreases in BA, G-Mean, and AUC, reaching 1.11, 2.11, and 0.77 percentage points, respectively, together with a F1-Score decrease of 0.99 percentage points. These results support the usefulness of hypergraph-guided synthesis. Removing DWT also decreases all four average scores, including the largest F1-Score decrease of 1.01 percentage points and a G-Mean decrease of 1.45 percentage points, supporting the contribution of wavelet-domain generation.

Removing quality pruning decreases the average scores across all four metrics by 0.40 to 0.65 percentage points, suggesting that filtering hyperedges before synthesis is beneficial. Replacing degree-based allocation with uniform allocation also lowers all four average scores, by 0.18 to 0.39 percentage points, supporting the use of vertex degrees to determine how many synthetic samples to generate from each minority-class sample. The details of the results are provided in \textbf{Appendix G of Supplementary Materials}. Overall, these results show that each component contributes to the effectiveness of FH-Sampling.

\subsection{Parameter Sensitivity Analysis}

We investigate the sensitivity of FH-Sampling to the neighborhood size $k$ and quality threshold $\theta$. Specifically, we evaluate the full grid of $k \in \{3,5,7,9\}$ and $\theta \in \{0.1,0.2,\ldots,0.9\}$. We compute $\sigma(k)$ as the sample standard deviation across feasible $k$ values at each fixed $\theta$, then average over valid $\theta$ settings. Similarly, $\sigma(\theta)$ averages the conditional sample standard deviations across feasible $\theta$ values at each fixed $k$.

\begin{table}[t]
\centering
\caption{Average parameter sensitivity on the 24 datasets}
\label{tab:parameter_sensitivity}
\begingroup
\footnotesize
\setlength{\tabcolsep}{6pt}
\renewcommand{\arraystretch}{1.15}
\resizebox{0.9\columnwidth}{!}{%
\begin{tabular}{@{}lcccc@{}}
\toprule[0.9pt]
Parameter & BA & F1-Score & G-Mean & AUC \\
\midrule[0.55pt]
$\sigma(k)$ & 0.0110 & 0.0090 & 0.0180 & 0.0073 \\
$\sigma(\theta)$ & 0.0072 & 0.0077 & 0.0084 & 0.0057 \\
\bottomrule[0.9pt]
\end{tabular}%
}
\endgroup
\end{table}

As shown in Table~\ref{tab:parameter_sensitivity}, the average values of $\sigma(k)$ range from 0.0073 to 0.0180, while those of $\sigma(\theta)$ range from 0.0057 to 0.0084. The details of the results are provided in \textbf{Appendix H of Supplementary Materials}. These results indicate relatively small average performance variations over the feasible settings, with lower average sensitivity to $\theta$ than to $k$ across all four metrics. The main experiments retain the fixed settings $k=5$ and $\theta=0.5$.

\subsection{Runtime Comparison}

We compare the sampling runtime of the 14 methods on the same 24 datasets. Runtime includes sampling and internal model fitting, but excludes classifier training, external data loading, result saving, and hyperparameter search. A theoretical analysis of the time complexity of FH-Sampling is provided in \textbf{Appendix A.2 of Supplementary Materials}.

\begin{figure}[t]
\centering
\includegraphics[width=1\columnwidth]{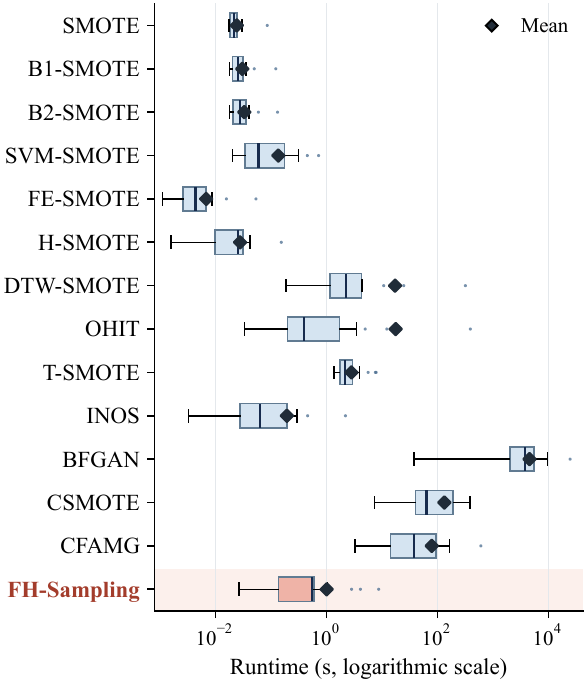}
\caption{Runtime distributions of the 14 sampling methods. Diamonds denote
means, boxes show quartiles, and inner lines mark medians.
Whiskers follow the $1.5\times\mathrm{IQR}$ rule; dots are outliers.}
\label{fig:runtime_comparison}
\end{figure}

As shown in Fig.~\ref{fig:runtime_comparison}, FH-Sampling achieves a mean runtime of 1.01s and a median runtime of 0.55s across the 24 datasets. Its mean runtime is lower than those of DTW-SMOTE (17.29s), OHIT (17.76s), T-SMOTE (2.82s), BFGAN (4576.13s), CSMOTE (133.66s), and CFAMG (79.18s). Although several simpler sampling methods are faster, FH-Sampling achieves stronger average classification performance while maintaining a relatively low sampling cost. These results indicate a favorable balance between performance and computational efficiency. The details of the results are provided in \textbf{Appendix I of Supplementary Materials}.

\section{Application to Aircraft Trajectory Generation}
\label{aircraft trajectory generation}

We further apply the proposed FH-Sampling method to the generation of high-maneuverability aircraft trajectories. We use a simulated flight dataset\footnote{https://github.com/Rossi0990/Simulation-data-of-high-maneuverability-aircraft.git} can be downloaded and compare FH-Sampling with TimeGAN~\cite{yoon2019time}, Diffusion-TS~\cite{yuan2024diffusion}, and TimeVAE~\cite{desai2021timevae}. 

We use the squared maximum mean discrepancy ($\mathrm{MMD}^{2}$)~\cite{gretton2012kernel} to measure the similarity between the distributions of generated and real trajectories. Lower values indicate greater similarity between the two distributions. We also use kernel density estimation (KDE)~\cite{yuan2024diffusion} to visualize the distributions of real and generated trajectory data. Greater overlap between the density curves indicates greater similarity in the distributions shown.


\begin{table}[t]
\centering
\caption{Generation quality on the simulated aircraft trajectory dataset}
\label{tab:aircraft_application}
\begingroup
\small
\setlength{\tabcolsep}{7pt}
\renewcommand{\arraystretch}{1.2}
\begin{tabular}{@{}lc@{}}
\toprule[0.9pt]
Method & $\mathrm{MMD}^{2}$ $(\times 10^{-3})$ $\downarrow$ \\
\midrule[0.55pt]
TimeGAN & 162.520\,\textpm\,2.213 \\
Diffusion-TS & 18.307\,\textpm\,1.685 \\
TimeVAE & \underline{7.945\,\textpm\,0.884} \\
\textbf{FH-Sampling} & \textbf{4.813\,\textpm\,0.351} \\
\bottomrule[0.9pt]
\end{tabular}
\endgroup
\end{table}

\begin{figure*}[t]
\centering
\includegraphics[width=\textwidth]{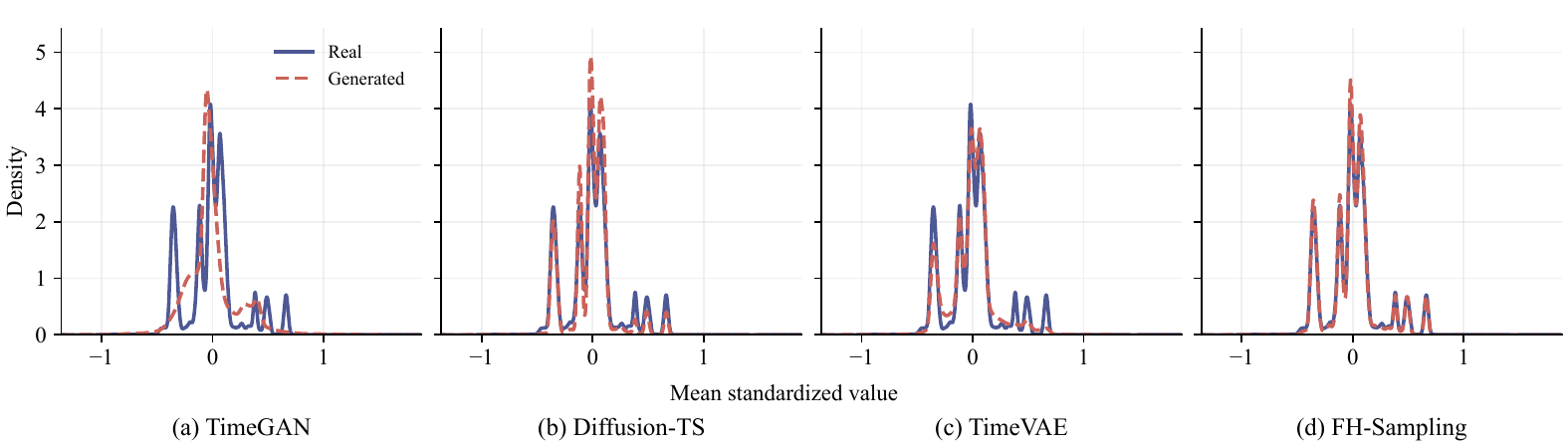}
\caption{Kernel density estimates for real and generated aircraft trajectory data.}
\label{fig:aircraft_generation_kde}
\end{figure*}

As shown in Table~\ref{tab:aircraft_application}, FH-Sampling achieves the lowest $\mathrm{MMD}^{2}$ of 0.004813, which is 39.4\% lower than that of TimeVAE, the second-best method. The density curves in Fig.~\ref{fig:aircraft_generation_kde} illustrate where the methods differ. TimeGAN smooths over several distinct peaks in the real-data curve, while Diffusion-TS and TimeVAE reproduce the main peaks but underestimate several smaller peaks at positive values. FH-Sampling more closely matches both the main peaks and these smaller peaks, capturing features of the real-data distribution. The lower $\mathrm{MMD}^{2}$ and greater overlap between the density curves show that FH-Sampling can generate aircraft trajectories whose distribution more closely matches that of the real flight data.

Further details on the dataset, baselines, and experimental settings, along with the downstream forecasting results, are provided in \textbf{Appendix J of Supplementary Materials}. 

\section{Conclusion}
\label{conclusion}

We propose FH-Sampling to address two limitations of interpolation-based oversampling for ITSC: the limited use of high-order relationships among samples and the loss of frequency information during synthesis. FH-Sampling constructs a hypergraph in the time domain to capture relationships among multiple samples. These relationships guide quality pruning and sample allocation to reduce the influence of noisy samples. The sample hypergraph guides synthesis in the wavelet domain, where the method varies local details while retaining the overall trend of the target sequence, thereby helping preserve frequency information that may be lost during direct interpolation in the time domain.

Experiments on imbalanced time-series datasets demonstrate that FH-Sampling improves the classification performance of various classifiers across different evaluation metrics, including BA, F1-Score, G-Mean, and AUC. The runtime comparison also shows that FH-Sampling achieves a favorable trade-off between classification performance and sampling time. The simulated aircraft trajectory generation study further demonstrates the potential of FH-Sampling for practical applications. In future work, we will evaluate the method on real-world multivariate datasets and investigate its use in a wider range of time-series applications.

\bibliographystyle{IEEEtran}
\bibliography{references}

\end{document}